\documentclass[letterpaper, 10 pt, conference]{ieeeconf}  

\IEEEoverridecommandlockouts                              

\usepackage{graphicx} 
\usepackage{epsfig} 
\usepackage{mathptmx} 
\usepackage{times} 
\usepackage{amsmath} 
\usepackage{amssymb}  

\usepackage{caption}
\usepackage{subcaption}
\usepackage{float}
\usepackage[export]{adjustbox}

\title{\LARGE \bf GarchingSim: An Autonomous Driving Simulator with Photorealistic Scenes and Minimalist Workflow
}

\author{Liguo Zhou*, Yinglei Song*, Yichao Gao*, Zhou Yu, Michael Sodamin, Hongshen Liu, Liang Ma, Lian Liu, \\ Hao Liu, Yang Liu, Haichuan Li, Guang Chen, Alois Knoll
\thanks{This work was supported by the Sino-German fond (5091331) and the China Scholarship Council (201806270244).}
\thanks{All authors are with the Chair of Robotics, Artificial Intelligence and Real-time Systems, Technical University of Munich, Parkring 13, 85748 Garching, Germany
        {\tt\small liguo.zhou@tum.de knoll@in.tum.de}}%
\thanks{* These authors contributed equally to this work.}
}

\begin{document}

\maketitle
\thispagestyle{empty}
\pagestyle{empty}

\begin{abstract}

Conducting real road testing for autonomous driving algorithms can be expensive and sometimes impractical, particularly for small startups and research institutes. Thus, simulation becomes an important method for evaluating these algorithms. However, the availability of free and open-source simulators is limited, and the installation and configuration process can be daunting for beginners and interdisciplinary researchers. We introduce an autonomous driving simulator with photorealistic scenes, meanwhile keeping a user-friendly workflow. The simulator is able to communicate with external algorithms through ROS2 or Socket.IO, making it compatible with existing software stacks. Furthermore, we implement a highly accurate vehicle dynamics model within the simulator to enhance the realism of the vehicle's physical effects. The simulator is able to serve various functions, including generating synthetic data and driving with machine learning-based algorithms. Moreover, we prioritize simplicity in the deployment process, ensuring that beginners find it approachable and user-friendly. 

\end{abstract}

\section{INTRODUCTION}

Comprehensive testing is an essential and necessary aspect of autonomous vehicle development. While real-world testing is a reliable method, achieving convincing validation results often requires millions of kilometers of driving, which makes it costly and time-consuming. The annual Autonomous Mileage Report \cite{disengagement} published by the California Department of Motor Vehicles shows that numerous manufacturers conducted over 10 million miles of testing in 2022. However, such extensive testing is impractical for small startups and research institutes. Additionally, real-world testing often struggles to cover a sufficient number of corner cases, as common traffic conditions tend to repeat. Investigating vehicle performance under extreme scenarios may require additional infrastructure costs. Given the wide range of parameters in urban traffic environments, including straight lines, corners, crosswalks, vehicles, cyclists, pedestrians, and various weather conditions \cite{son2019simulation}, it becomes impossible to cover all combinations on the road or in a test field. As a result, simulation plays a crucial role in bridging the gap between algorithm development and real-world testing. It allows researchers to create diverse scenarios in virtual environments and conduct safety-critical tests, such as emergency brakes and automated overtaking on highways, without practical risks.

Ideally, a simulator should provide platforms for various functionalities, including perception, localization and mapping, path planning, and control. This entails meeting several requirements. Firstly, to accurately replicate real-world features, the simulated environment must include high-quality 3D objects such as buildings, roads, and vehicles. Moreover, precise vehicle physics simulation is crucial for an authentic experience. As autonomous driving usually demands diverse sensors, a flexible sensor suite should be available. With the advancements in deep learning, having the ability to record training data is beneficial for algorithm development. Additionally, a well-structured traffic system, encompassing traffic lights, traffic signs, and other traffic agents, allows researchers to test autonomous driving functions under various traffic scenarios, including those that are unlikely to occur in real life. Lastly, the simulator should offer interfaces to external software, enabling the direct usage of existing software packages for tasks like detection or path planning.

In this paper, based on Unity 3D game engine~\cite{unity}, we present a new autonomous driving simulator with photorealistic scenes and physically precise objects. The 3D environment is designed based on real-world cities. The 3D models and textures are created by a group of artists specializing in game engineering and design. Furthermore, we manually integrate traffic signs and implemented control scripts for traffic lights. The movement of other traffic agents follows the waypoint-based traffic flow. We have incorporated various sensors, including cameras, radar, inertial measurement units (IMU), and global navigation satellite systems (GNSS). The Unity perception camera allows for the recording of ground truths for object detection or semantic segmentation during driving. To accurately replicate dynamic behaviors, we implement a vehicle dynamics model similar to that in a daily car. Communication with external software, such as PyTorch~\cite{pytorch} and TensorFlow~\cite{tensorflow}, is done through either Socket.IO or ROS2. They both enable the system to operate within a server multi-client architecture, offering flexibility and scalability. Finally, we prioritize ease of deployment, emphasizing a ``plug-and-play" style for the platform. The source code is released at ``https://github.com/tum-autonomousdriving/autonomous-driving-simulator".

The rest of this paper is structured as follows: Section \uppercase\expandafter{\romannumeral2} presents a review of previous simulation-related technologies in the autonomous driving field. In section \uppercase\expandafter{\romannumeral3}, we depict the features and capabilities of our simulator. Section \uppercase\expandafter{\romannumeral4} provides examples of various applications and algorithms that have been tested within its environment. Section \uppercase\expandafter{\romannumeral5} forms a conclusion and outlines potential avenues for future development.

\begin{figure}
    \centering
    \includegraphics[width = 0.49\textwidth]{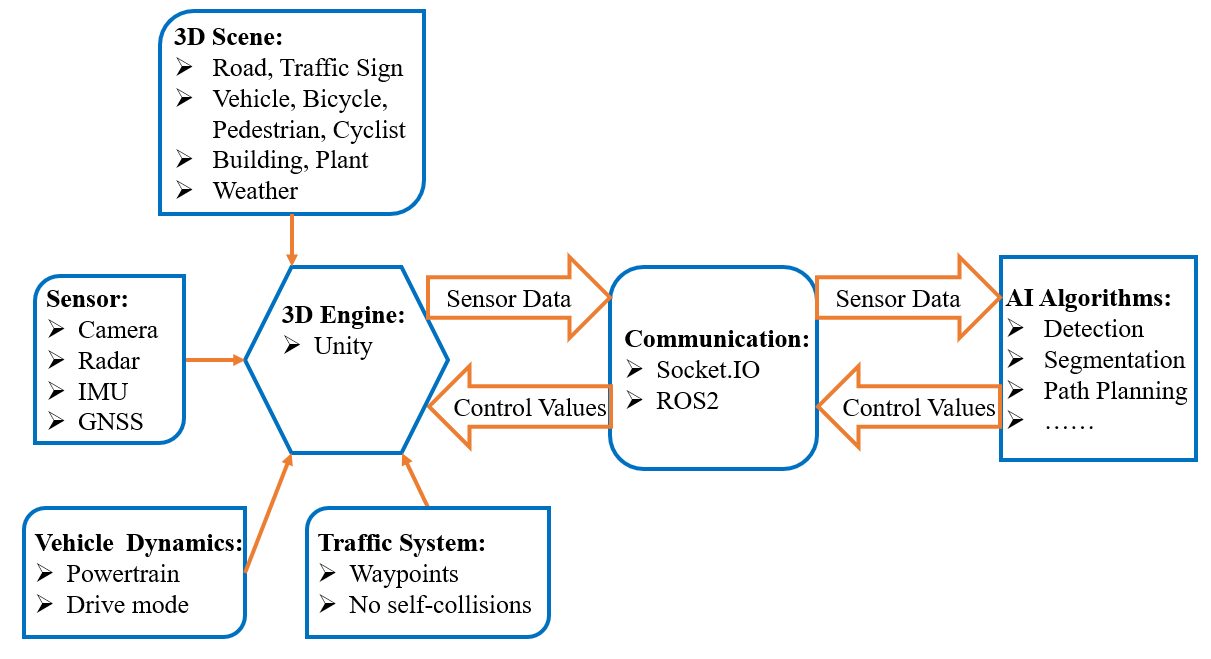}
    \caption{The framework of our autonomous driving simulator.}
    \label{fig: framework}
\end{figure}

\begin{figure*}[ht]
    \centering

    \begin{subfigure}[c]{0.49\textwidth}
        \includegraphics[width=\textwidth, height=5cm]{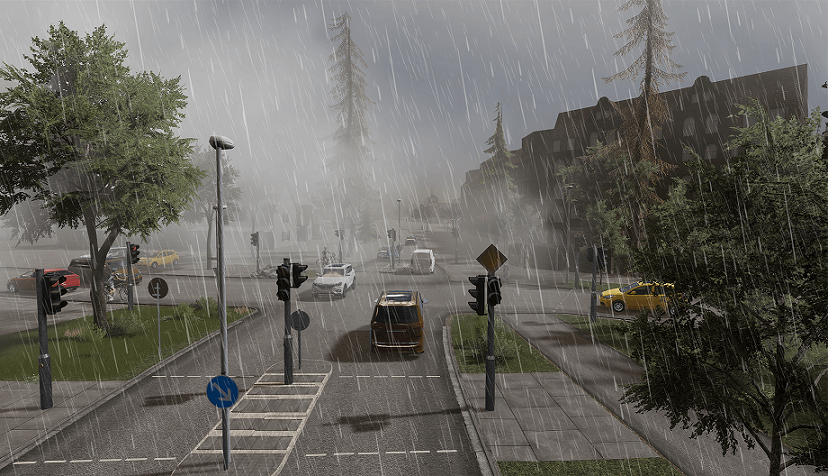}
        \caption{Rainy.}
    \end{subfigure}
    \begin{subfigure}[c]{0.49\textwidth}
        \includegraphics[width=\textwidth, height=5cm]{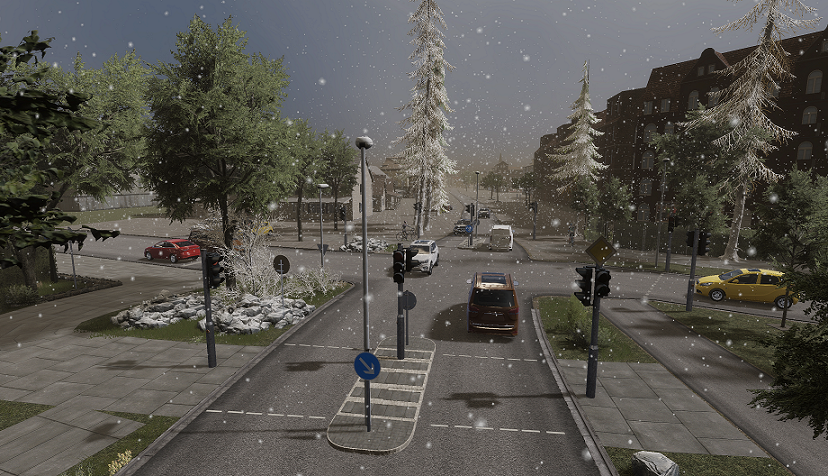}
        \caption{Snowy.}
    \end{subfigure} 
    \begin{subfigure}[c]{0.49\textwidth}
        \includegraphics[width=\textwidth, height=5cm]{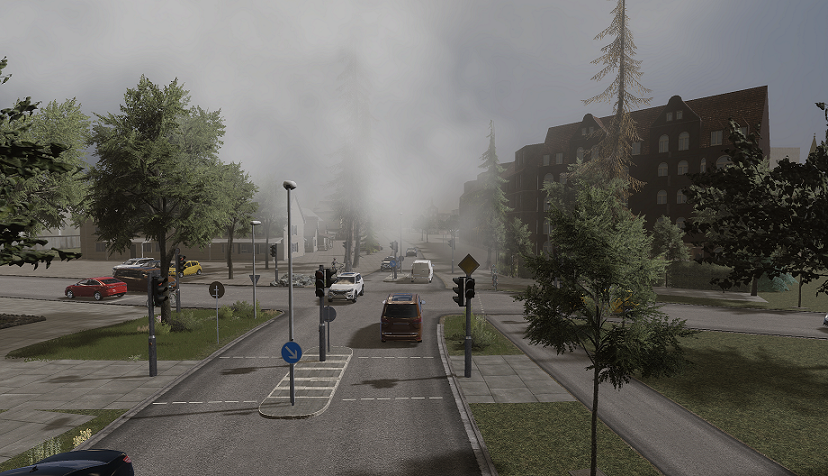}
        \caption{Foggy.}
    \end{subfigure}
    \begin{subfigure}[c]{0.49\textwidth}
        \includegraphics[width=\textwidth, height=5cm]{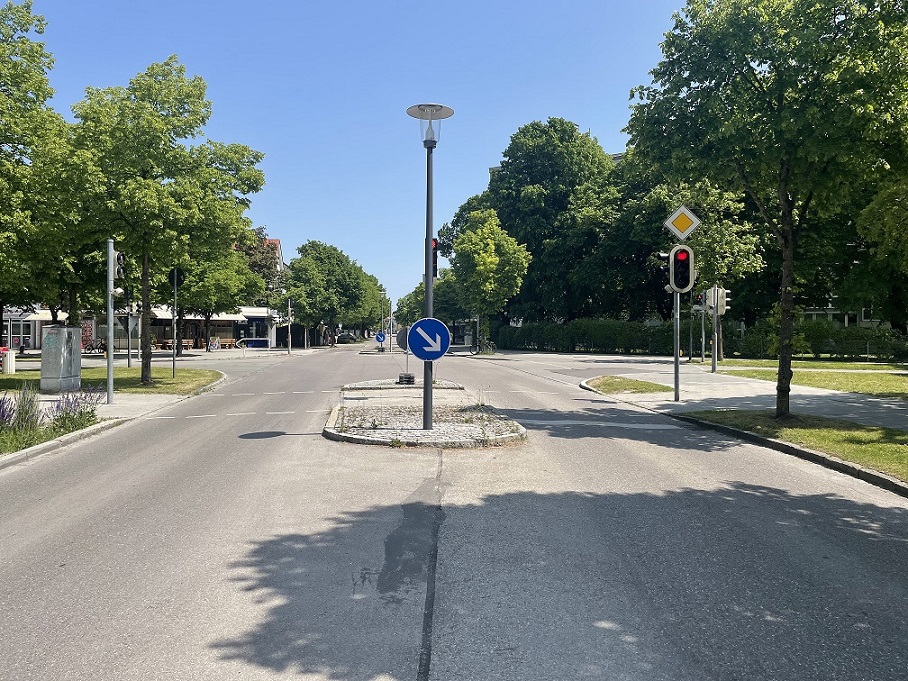}
        \caption{Real world scene.}
    \end{subfigure} 
    \caption{Simulation environment under different weathers and real-world scene}
    \label{fig: weather}
\end{figure*}

\section{Related Work}
Simulation technology~\cite{nrp} plays a crucial role in the development of vehicles. Initially, the focus was primarily on vehicle dynamics simulation. Software like Carsim ~\cite{carsim} gained significant popularity as a widely used tool in this domain. In recent versions, Carsim has also incorporated modules for advanced driver assistance systems (ADAS) and autonomous vehicle simulation. However, it lacks support for common sensors such as cameras or radar. Instead, it relies on a dedicated ADAS sensor to measure the range between the ego vehicle and objects.  Compared to other simulators, Carsim's 3D environment still falls short in terms of its fidelity to the real world.

Gazebo \cite{koenig2004design} is a popular simulation platform within the field of robotics. It is typically used together with Robot Operating System (ROS) \cite{qian2014manipulation} \cite{takaya2016simulation}. Gazebo's physics engine offers a level of accuracy that surpasses traditional game engines. However, creating large-scale 3D scenarios in Gazebo can be challenging and time-consuming. Objects in Gazebo are primarily created through code, lacking the more intuitive graphical user interface found in other tools. In terms of visual rendering, Gazebo's capabilities do not match those of professional game engines such as Unity \cite{unity} and Unreal Engine \cite{unreal}.

Big companies have built their own simulators like Carcraft \cite{fadaie2019state} used by Waymo, and The Matrix \cite{wiggers2020challenges} used by Cruise. These simulators are restricted to internal use and are not publicly available. Within the autonomous driving community, there are also several dedicated open-source simulators available. CARLA \cite{carla} is the first simulator to incorporate autonomous driving functions. It is built upon the Unreal Engine \cite{unreal}, and its 3D environment is created from scratch by a team of digital artists. CARLA currently supports a range of sensors, including cameras, LiDAR, radar, IMU, and GNSS. However, due to its large-scale nature, the process of installing the software and configuring the necessary setups can be time-consuming. Many beginners have also reported encountering some initial difficulties when getting started with CARLA.

Another open-source simulator is LGSVL \cite{rong2020lgsvl}, which utilizes the Unity game engine as its foundation. Unlike CARLA, which has various functionalities built inside the simulator, LGSVL is designed to connect with external autonomous driving (AD) stacks such as Autoware \cite{autoware} and Baidu Apollo \cite{apollo}. This connection is facilitated through different bridges, including ROS1/ROS2 bridge and Cyber RT, a custom bridge for Apollo. A notable feature of LGSVL is its ability to create, edit, and export HD maps based on existing 3D environments. Unfortunately, LG has decided to suspend active development of the SVL simulator, effective January 1, 2022. Consequently, no further updates or improvements to the simulator will be released.

There are also commercial autonomous driving simulators available on the market. For example, MATLAB/Simulink \cite{matlab} by Mathworks. The Automated Driving Toolbox™ of MATLAB offers tools for designing, simulating, and testing ADAS and autonomous driving systems. One key advantage of MATLAB is its code generation capability, allowing for faster prototyping by generating C/C++ code. Another commercial simulator is Prescan \cite{prescan}, developed by Siemens. Prescan enables users to quickly replicate real-world traffic scenarios using elements from its database. It also supports hardware-in-the-loop (HIL) simulation, a common practice for evaluating electronic control units (ECUs). Nevertheless, the 3D environments created in these two simulators do not provide a photo-realistic experience. Additionally, as they are not open-source, customizing the settings and functionalities may pose some challenges.

\section{Features}

\subsection{Software Framework}
Our simulator is developed using the Unity game engine \cite{unity}. We implement a design where the simulated environments and algorithms run independently. As shown in Fig. \ref{fig: framework}, traffic system and sensor models operate within the simulator. Along with 3D scene and vehicle dynamics models, the simulator is able to present real traffic scenarios effectively. Communication between the simulator and algorithms is facilitated through either Socket.IO or ROS2. The concept behind this architecture is to transmit sensor readings from the simulator to the algorithms via the communication interface. These sensor readings serve as inputs for the algorithms, while the algorithms, in turn, send back control signals such as throttle and steering angle. This system architecture is particularly beneficial for users who require simulations to run on multiple computers. With Socket.IO's server-client connection, the main simulation environment can run on a powerful desktop equipped with a GPU, while multiple laptops can control individual ego vehicles.
Regarding ROS2, it naturally supports a master-slave architecture, enabling seamless communication between the master and slave machines through a publisher-subscriber network. This publisher-subscriber structure enables functionalities such as vehicle-to-vehicle (V2V) and vehicle-to-everything (V2X) interactions.

\subsection{HD Object and Building}
We implement the High Definition Render Pipeline (HDRP) of Unity to achieve a photo-realistic scene in our simulator. For the simulated environment, we choose some real-world cities around our workplace. The buildings, roads, and crosswalks in the simulator closely resemble the actual map of real cities. Our objective is to create test environments that mirror the real world, enabling us to evaluate the sim-to-real quality of the algorithms accurately. To achieve this, a dedicated group of artists specializing in game engineering and media design meticulously create 3D objects from scratch. These objects are initially crafted with high-precision surfaces in 3ds Max and subsequently imported into the Unity scene editor. Through this process, we ensure that the 3D objects in the simulator maintain a high level of fidelity to their real-world counterparts. Furthermore, we provide full customization options for weather and illumination conditions within the simulator. Fig. \ref{fig: weather} demonstrates the simulation environment under different weather as well as a real-world scene. Users can tailor these parameters according to their specific requirements.

\subsection{Vehicle Dynamics}
We use a specialized vehicle dynamics model in our simulator, allowing for the precise representation of realistic car behaviors. This vehicle model encloses both internal combustion engines (ICE) and electric drives, and various drive modes are available such as front-wheel drive, rear-wheel drive, and full-wheel drive. Each drive mode accurately captures the distinct dynamics of different vehicle setups, enabling comprehensive testing to ensure algorithmic safety across diverse scenarios. To ensure accurate control over dynamics parameters, we implement Proportional-Integral-Derivative (PID) controller. This controller is specifically designed to facilitate rapid convergence to the desired setpoint, enabling efficient and effective control of the simulated vehicle's steering and speed.

\begin{figure}
    \centering
    \includegraphics[width = 0.49\textwidth]{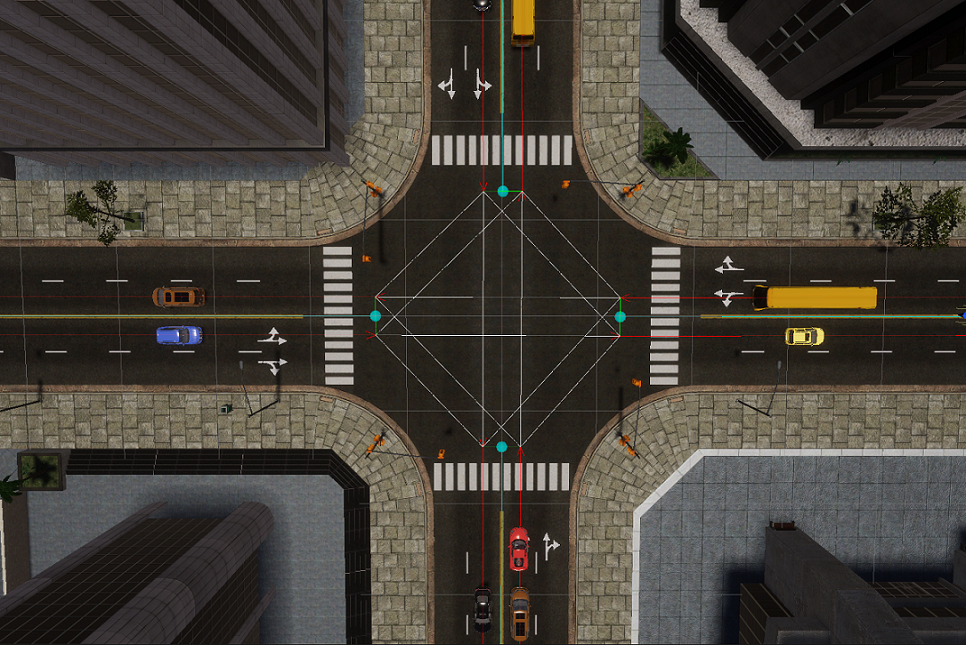}
    \caption{Waypoint-based traffic system}
    \label{fig: traffic}
\end{figure}

\begin{figure*}
    \centering

    \begin{subfigure}[c]{0.32\textwidth}
        \includegraphics[width=\textwidth, height=3.5cm]{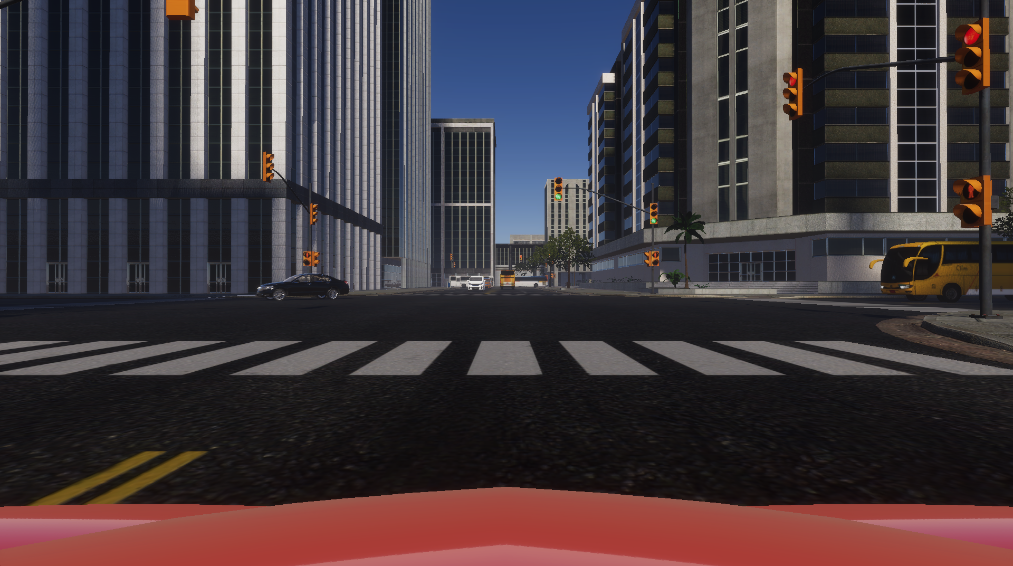}
        \caption{RGB camera.}
    \end{subfigure}
    \begin{subfigure}[c]{0.32\textwidth}
        \includegraphics[width=\textwidth, height=3.5cm]{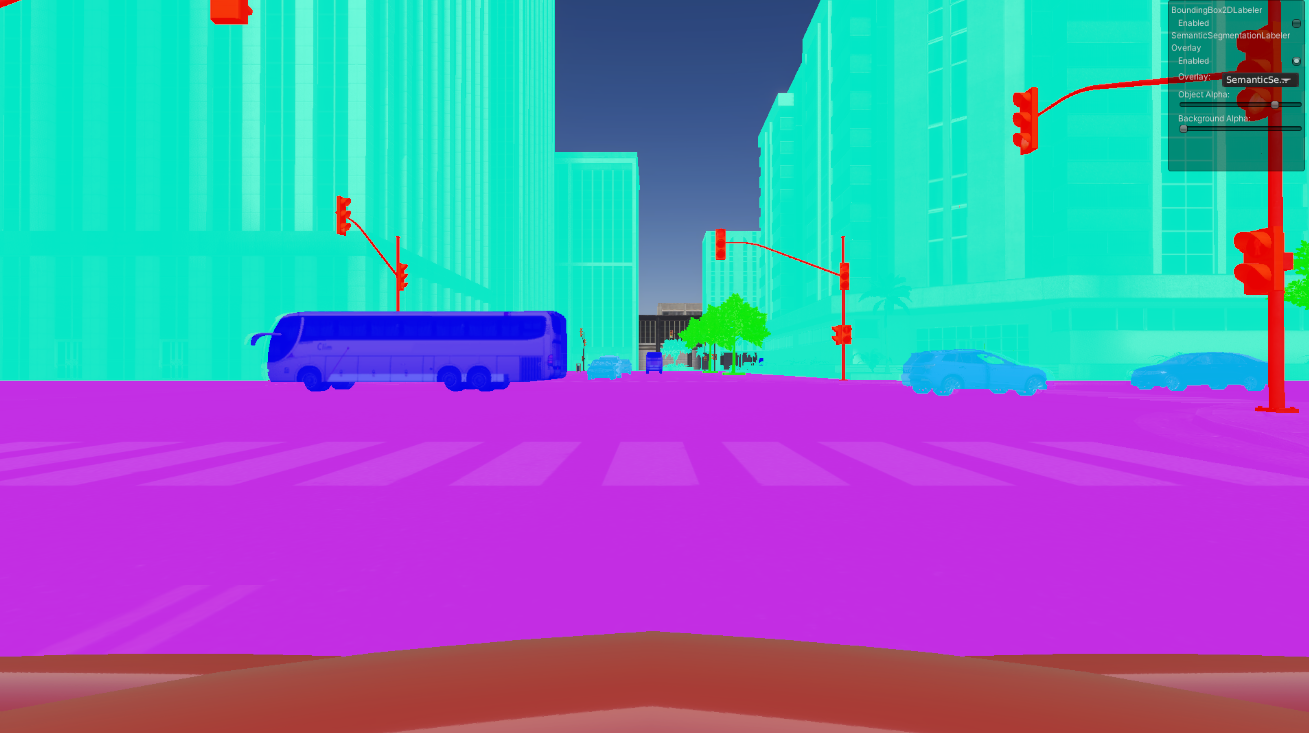}
        \caption{Semantic camera.}
    \end{subfigure} 
    \begin{subfigure}[c]{0.32\textwidth}
        \includegraphics[width=\textwidth, height=3.5cm]{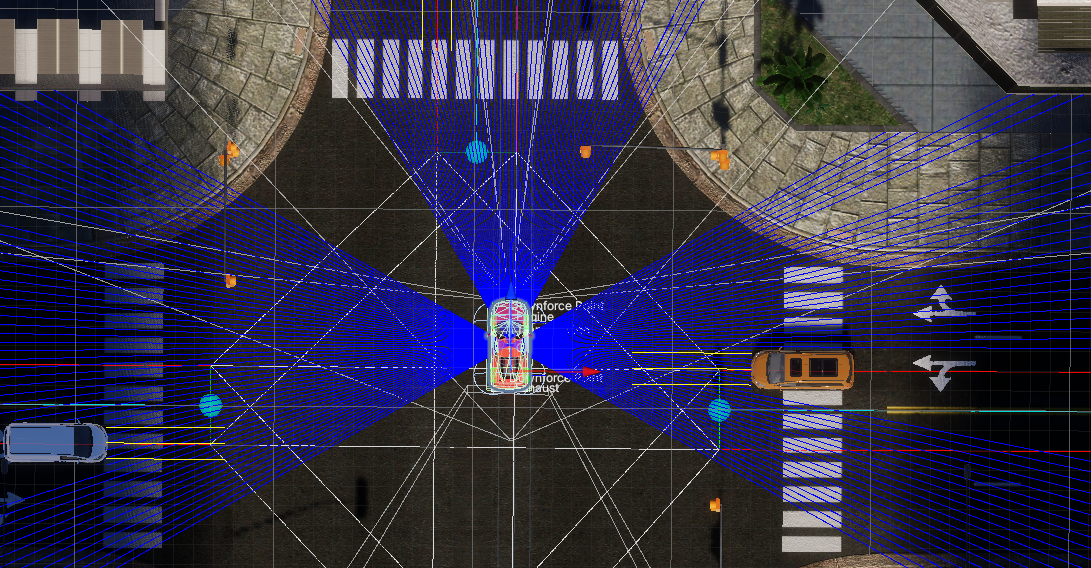}
        \caption{Radar.}
    \end{subfigure}

    \caption{RGB camera, semantic camera, and radar}
    \label{fig: sensors}
\end{figure*}

\subsection{Traffic System}
In the traffic simulation, we manually define waypoints that govern the behavior of other traffic agents. An example of waypoints on the intersection is shown in Fig. \ref{fig: traffic}. This approach is particularly suitable for low-performance computers. Basically, the simulated vehicles adhere to these predetermined routes while also possessing the capability to detect and respond to traffic lights. When the lights transition to red, the vehicles will halt accordingly. Additionally, we include an elementary collision avoidance system. Each vehicle is equipped with a forward-facing ray that detects potential obstacles. When the ray intersects with objects, the system calculates the length of the ray and returns this value to the relevant function. If the measured distance falls below a predefined threshold, the vehicle will automatically come to a stop, waiting for the critical situation to disappear. In the extreme case that the collision does happen, the involved vehicles are promptly relocated away from their current positions and regenerated elsewhere within the simulated environment. This guarantees that the ego vehicle remains unaffected by accidents caused by other traffic agents and potential traffic congestion is averted.

\subsection{Sensors}
Our simulator supports various sensors. Currently, the supported sensors include cameras, radar, IMU, and GNSS. Each of these sensors can be fully customized to meet specific requirements. Fig. \ref{fig: sensors} depicts visualizations of some sensors. The configuration parameters, such as ROS2 topics, Socket.IO IPs, publishing rate, reference frames, as well as sensor locations and orientations, can all be conveniently adjusted via a YAML file.
\subsubsection{Cameras}
The camera module provides multiple camera modes, including RGB camera, semantic segmentation camera, and instance segmentation camera. Furthermore, the intrinsic and extrinsic parameters of the camera can be easily modified by the users.
\subsubsection{Radar}
We develop a radar system capable of accurately measuring the distances to objects within specified sector areas. By default, the simulator includes three radars positioned at the front, left, and right sides of the ego vehicle
\subsubsection{IMU}
To facilitate tasks such as sensor fusion, our ego vehicle model is equipped with an IMU that reports real-time information on acceleration and angular velocity.
\subsubsection{GNSS}
The global position of the ego vehicle can be easily accessed through the Unity Game Object properties.

\subsection{Communication between Simulator and Algorithms}
\subsubsection{Socket.IO}
We utilize Socket.IO to establish a connection between the simulator server and various clients or modules. It enables bidirectional and event-based communication, thus, is ideal for transmitting data and instructions between the simulator and other components of an autonomous driving system. The simulator server can emit events or data updates, which can then be received by the connected clients or modules. Similarly, a client can do it in a vice-versa way. This communication mechanism allows for seamless integration of different components. For example, the simulator server can send sensor data to a perception module, and receive control commands from a planning module. This architecture is also suitable for controlling multiple ego vehicles in the simulated environment. The real-time capabilities of Socket.IO facilitate the efficiency of communication and reduce latency.

\subsubsection{ROS 2}
We implement an open-source communication solution between Unity and ROS2 middleware \cite{ros2unity}. Compared to traditional bridging methods, this non-bridge solution has a higher performance and considerably lower latency. It uses the ROS2 middleware stack (rcl layer and below), which means ROS2 nodes are directly running in the simulation. As a result, ROS2 messages can be transmitted to their corresponding topics and effortlessly received by nodes operating outside of the simulator. The communication solution provides flexibility for customizing published messages through scripts. Currently, the communication module focuses on transmitting sensor readings while receiving steering, acceleration, and braking commands. This enables seamless integration with external autonomous driving software stacks, such as Autoware.  Additionally, the ROS2 interface facilitates easy integration with existing ROS2 packages like image\_pipeline and move\_base, providing researchers with the ability to leverage these packages within the simulation environment.

\subsection{Easy Deployment and Portability}
The simplified version of our simulator is provided as an executable Unity file, which includes predefined ROS2 topics and Socket.IO interfaces. By default, the simulator is equipped with RGB cameras, radar, IMU, and GNSS sensors. This version is primarily designed for educational purposes, such as practical courses for students.

In addition to the simplified version, we also offer a developer version of the simulator, which provides extensive customization options. Users can tailor sensor combinations, vehicle models, buildings, and other traffic agents according to their specific research requirements. This highly customizable version is particularly well-suited for research purposes, allowing researchers to evaluate the performance of algorithms related to autonomous driving in a controlled and customizable environment.

The Unity file is compatible with both Windows and Linux systems, ensuring cross-platform usability. ROS2 and Socket.IO also support these two platforms.

\begin{figure}
    \centering
    \includegraphics[width = 0.49\textwidth]{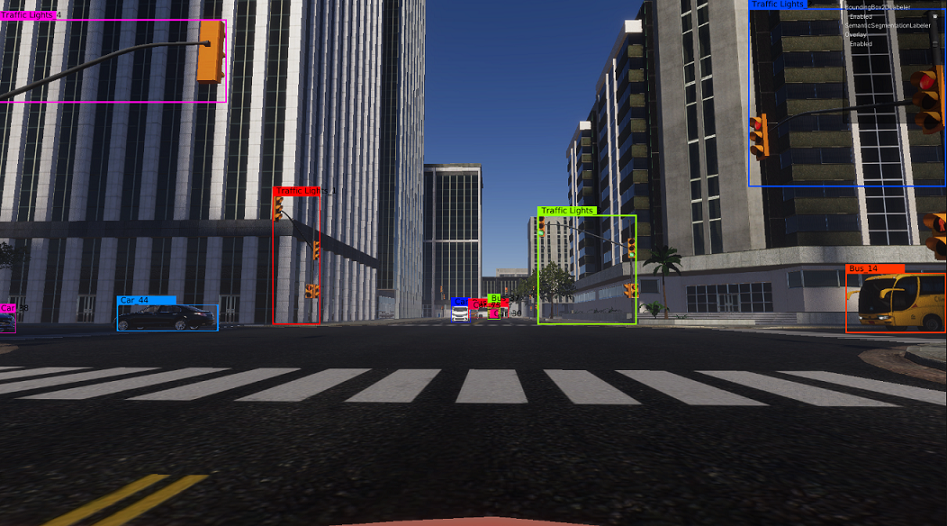}
    \caption{Automatic labeling}
    \label{fig: labeling}
\end{figure}

\section{Applications}
\subsection{Generating Synthetic Training Data}
Recent advances in deep learning have led to great improvements in the field of autonomous driving. However, the process of collecting real-world training data can be both time-consuming and expensive. To address this challenge, our simulator incorporates a dedicated mode for generating synthetic training data. In this mode, users can specify the types of objects they want to detect or segment within the simulator. For object detection tasks, our simulator automatically generates ground truth 2D/3D bounding boxes for each selected object, as shown in Fig. \ref{fig: labeling}. Similarly, for segmentation tasks, distinct color masks are applied to the objects. As the 3D models of these objects are known in the simulator, the generated bounding boxes and color masks are more precise than those obtained through manual human labeling. The labels, origin, and dimension of the bounding box to each object will be saved in a JSON file under the local folder.

\subsection{Imitation Learning}
Imitation learning (IL) is an intelligent learning strategy that leverages expert trajectories to guide decision-making and control\cite{attia2018global}. Expert trajectories consist of state-action pairs, which are extracted to create a dataset. IL aims to learn the underlying relationships between states (representing features) and actions (displaying labels), with the objective of maximizing the agent’s ability to replicate expert trajectories. We use this autonomous driving simulator to adequately provide a stable validation environment for large-scale imitation learning strategies.

\begin{figure}
    \centering
    \includegraphics[width = 0.49\textwidth]{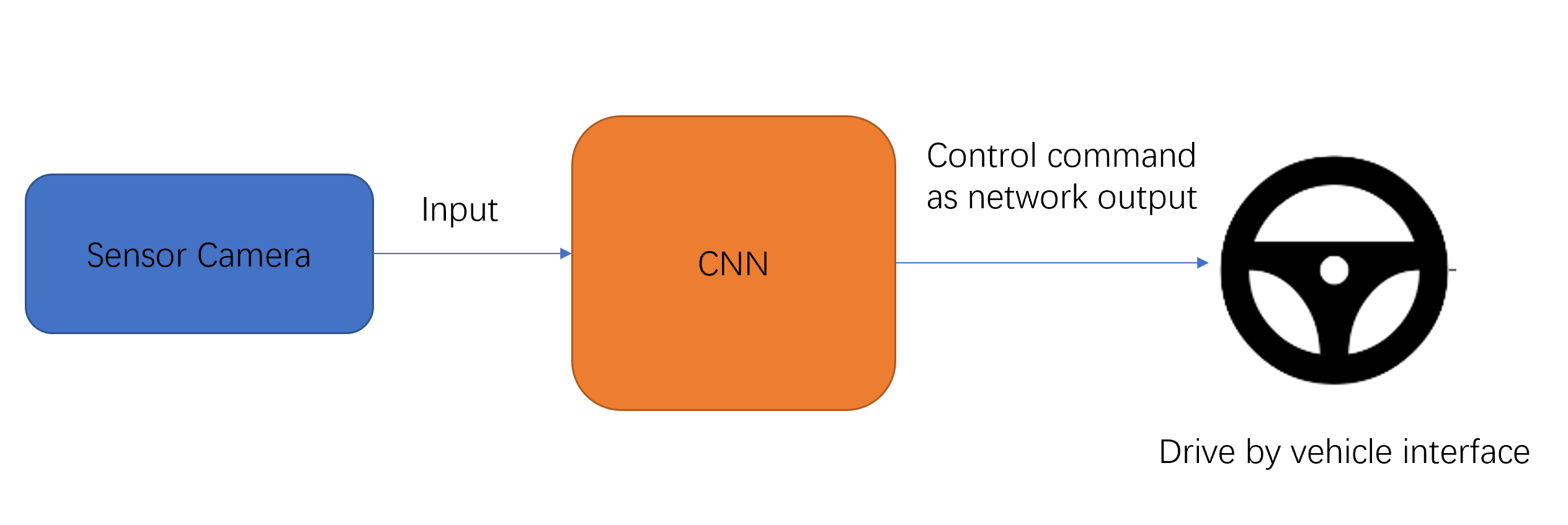}
    \caption{Imitation learning workflow in the simulator}
    \label{fig: IL}
\end{figure}

The current dominant frameworks for imitation learning are based on an end-to-end learning approach \cite{bojarski2016end}. That is, the real-time sensor information is input, and a core model processes this input to directly generate vehicle control commands. However, validating such end-to-end approaches directly in real-world scenarios is challenging due to associated risks and costs. A simulated environment is more suitable for this validation task. The virtual vehicle in our simulator can be tested in different scenarios and on different roads without discrimination. Fig. \ref{fig: IL} illustrates the workflow our simulator provides for imitation learning applications.

We can artificially set the expert behavior by driving the vehicle through keyboard commands. In the simulator, the vehicle is equipped with cameras, which receive a live feed from the simulated environment and transmit it to the client. The trained convolutional neural network (CNN) in the client processes the live input and simultaneously generates a control signal that is sent back to the simulator via the Socket.IO. Subsequently, the vehicle receives the control commands and implements real-time control.

\subsection{Reinforcement Learning}
In addition to imitation learning, reinforcement learning methods \cite{sallab2017deep} have gained attention for training autonomous driving algorithms in an end-to-end style. Reinforcement learning is often considered a preferred choice over imitation learning due to its ability to handle complex environments, adapt to changes, and manage multi-objective tasks more effectively \cite{kiran2021deep}.

Our autonomous driving simulator can be used to generate large volumes of training data, including sensor readings, environmental information, and vehicle status. This data can be utilized to train reinforcement learning models, enabling the agent to acquire driving skills and decision-making strategies. The simulator automatically generates observation information from the camera sensor based on the current environment. Each observation $o_t \in \mathcal{O}$ consists of four sets of tensors from the previous timesteps, namely $o_t=\left\{[C, R, V, N]\right\}$. The $C$ tensor represents a concatenated $640 \times 320n \times 3$ camera image, comprising $n$ number of $640 \times 320 \times 3$ RGB camera sensor images placed at different locations on the vehicle. The $R$ tensor denotes road features, $V$ captures vehicle features, and $N$ is a vector containing navigational features.
The simulator has embedded a wide range of virtual driving environments. This includes city roads, weather conditions, illuminations, motorways, intersections, pedestrians, car parks, traffic flow, etc. By constructing virtual environments in the simulator, we provide users with a convenient means to test and evaluate the performance of reinforcement learning algorithms in various driving scenarios.

\subsection{V2V and V2X}
A comprehensive autonomous system should not only rely on its own sensor reading but also consider the shared information from other intelligent agents. Acquiring precise positions of surrounding traffic agents, rather than estimating them solely from sensor data, will enhance the accuracy and effectiveness of the ego vehicle's decision-making process. The vehicle-to-vehicle (V2V) architecture is designed to facilitate the exchange of information among different vehicles. As other traffic participants, and even infrastructures, become part of this communication scheme, the system evolves into a vehicle-to-everything (V2X) framework. However, constructing those infrastructures is usually costly and challenging in the real world. Therefore the evaluation and testing are more viable to be conducted in a simulated environment \cite{ye2019deep}. By utilizing the node-topic architecture of ROS2, our simulator can efficiently distribute information among all intelligent agents.

\section{CONCLUSIONS}

We introduce an autonomous driving simulator named after our workplace Garching, that is built on the Unity game engine. By incorporating high-definition objects, buildings, and precise vehicle physics models, the simulator is able to reproduce real traffic scenarios and the dynamic behaviors of a car. GarchingSim provides flexible sensor suites so that users can customize them according to their preferences. The simulator serves as an ideal platform for evaluating machine learning-based algorithms. The communication between the simulator and external software is done via ROS2 interfaces or Socket.IO, enabling distributed simulation across multiple computers. Additionally, the existing autonomous driving software stacks or packages can be integrated into our system. We design GarchingSim as an open-source platform and remain committed to enhancing it by adding more features upon the feedback of the users.

\addtolength{\textheight}{-12cm}   


\bibliographystyle{IEEEtran} 
\bibliography{ref}

\begin{thebibliography}{10}
\providecommand{\url}[1]{#1}
\csname url@samestyle\endcsname
\providecommand{\newblock}{\relax}
\providecommand{\bibinfo}[2]{#2}
\providecommand{\BIBentrySTDinterwordspacing}{\spaceskip=0pt\relax}
\providecommand{\BIBentryALTinterwordstretchfactor}{4}
\providecommand{\BIBentryALTinterwordspacing}{\spaceskip=\fontdimen2\font plus
\BIBentryALTinterwordstretchfactor\fontdimen3\font minus \fontdimen4\font\relax}
\providecommand{\BIBforeignlanguage}[2]{{%
\expandafter\ifx\csname l@#1\endcsname\relax
\typeout{** WARNING: IEEEtran.bst: No hyphenation pattern has been}%
\typeout{** loaded for the language `#1'. Using the pattern for}%
\typeout{** the default language instead.}%
\else
\language=\csname l@#1\endcsname
\fi
#2}}
\providecommand{\BIBdecl}{\relax}
\BIBdecl

\bibitem{disengagement}
\BIBentryALTinterwordspacing
{California Department of Motor Vehicles}, ``Disengagement reports,'' 2022. [Online]. Available: \url{https://www.dmv.ca.gov/portal/vehicle-industry-services/autonomous-vehicles/disengagement-reports/}
\BIBentrySTDinterwordspacing

\bibitem{son2019simulation}
T.~D. Son, A.~Bhave, and H.~Van~der Auweraer, ``Simulation-based testing framework for autonomous driving development,'' in \emph{2019 IEEE International Conference on Mechatronics (ICM)}, vol.~1.\hskip 1em plus 0.5em minus 0.4em\relax IEEE, 2019, pp. 576--583.

\bibitem{unity}
\BIBentryALTinterwordspacing
{Unity Technologies}, ``{Unity Engine}.'' [Online]. Available: \url{https://www.unity.com/}
\BIBentrySTDinterwordspacing

\bibitem{pytorch}
A.~Paszke, S.~Gross, F.~Massa, A.~Lerer, J.~Bradbury, G.~Chanan, T.~Killeen, Z.~Lin, N.~Gimelshein, L.~Antiga \emph{et~al.}, ``Pytorch: An imperative style, high-performance deep learning library,'' \emph{Advances in neural information processing systems}, vol.~32, 2019.

\bibitem{tensorflow}
M.~Abadi, P.~Barham, J.~Chen, Z.~Chen, A.~Davis, J.~Dean, M.~Devin, S.~Ghemawat, G.~Irving, M.~Isard \emph{et~al.}, ``Tensorflow: a system for large-scale machine learning.'' in \emph{Osdi}, vol.~16, no. 2016.\hskip 1em plus 0.5em minus 0.4em\relax Savannah, GA, USA, 2016, pp. 265--283.

\bibitem{nrp}
W.~Cao, L.~Zhou, Y.~Huang, and A.~Knoll, ``Autonomous driving simulator based on neurorobotics platform,'' \emph{arXiv preprint arXiv:2301.00089}, 2022.

\bibitem{carsim}
\BIBentryALTinterwordspacing
{Mechanical Simulation Corporation}, ``{Carsim}.'' [Online]. Available: \url{https://www.carsim.com/}
\BIBentrySTDinterwordspacing

\bibitem{koenig2004design}
N.~Koenig and A.~Howard, ``Design and use paradigms for gazebo, an open-source multi-robot simulator,'' in \emph{2004 IEEE/RSJ International Conference on Intelligent Robots and Systems (IROS)(IEEE Cat. No. 04CH37566)}, vol.~3.\hskip 1em plus 0.5em minus 0.4em\relax IEEE, 2004, pp. 2149--2154.

\bibitem{qian2014manipulation}
W.~Qian, Z.~Xia, J.~Xiong, Y.~Gan, Y.~Guo, S.~Weng, H.~Deng, Y.~Hu, and J.~Zhang, ``Manipulation task simulation using ros and gazebo,'' in \emph{2014 IEEE International Conference on Robotics and Biomimetics (ROBIO 2014)}.\hskip 1em plus 0.5em minus 0.4em\relax IEEE, 2014, pp. 2594--2598.

\bibitem{takaya2016simulation}
K.~Takaya, T.~Asai, V.~Kroumov, and F.~Smarandache, ``Simulation environment for mobile robots testing using ros and gazebo,'' in \emph{2016 20th International Conference on System Theory, Control and Computing (ICSTCC)}.\hskip 1em plus 0.5em minus 0.4em\relax IEEE, 2016, pp. 96--101.

\bibitem{unreal}
\BIBentryALTinterwordspacing
{Epic Games, Inc.}, ``Unreal engine.'' [Online]. Available: \url{https://www.unrealengine.com/}
\BIBentrySTDinterwordspacing

\bibitem{fadaie2019state}
J.~Fadaie, ``The state of modeling, simulation, and data utilization within industry: An autonomous vehicles perspective,'' \emph{arXiv preprint arXiv:1910.06075}, 2019.

\bibitem{wiggers2020challenges}
K.~Wiggers, ``The challenges of developing autonomous vehicles during a pandemic,'' \emph{Venture Beat}, 2020.

\bibitem{carla}
A.~Dosovitskiy, G.~Ros, F.~Codevilla, A.~Lopez, and V.~Koltun, ``{CARLA}: {An} open urban driving simulator,'' in \emph{Proceedings of the 1st Annual Conference on Robot Learning}, 2017, pp. 1--16.

\bibitem{rong2020lgsvl}
G.~Rong, B.~H. Shin, H.~Tabatabaee, Q.~Lu, S.~Lemke, M.~Mo{\v{z}}eiko, E.~Boise, G.~Uhm, M.~Gerow, S.~Mehta \emph{et~al.}, ``Lgsvl simulator: A high fidelity simulator for autonomous driving,'' in \emph{2020 IEEE 23rd International conference on intelligent transportation systems (ITSC)}.\hskip 1em plus 0.5em minus 0.4em\relax IEEE, 2020, pp. 1--6.

\bibitem{autoware}
\BIBentryALTinterwordspacing
{Autoware Foundation}, ``Autoware.'' [Online]. Available: \url{https://github.com/autowarefoundation/autoware}
\BIBentrySTDinterwordspacing

\bibitem{apollo}
\BIBentryALTinterwordspacing
Baidu, ``Baidu apollo.'' [Online]. Available: \url{https://github.com/ApolloAuto/apollo}
\BIBentrySTDinterwordspacing

\bibitem{matlab}
\BIBentryALTinterwordspacing
MathWorks, ``{MATLAB/Simulink}.'' [Online]. Available: \url{https://www.mathworks.com/products/automated-driving.html}
\BIBentrySTDinterwordspacing

\bibitem{prescan}
\BIBentryALTinterwordspacing
Siemens, ``Prescan.'' [Online]. Available: \url{https://plm.sw.siemens.com/en-US/simcenter/autonomous-vehicle-solutions/prescan/}
\BIBentrySTDinterwordspacing

\bibitem{ros2unity}
\BIBentryALTinterwordspacing
{Robotec.AI}, ``Ros2 for unity.'' [Online]. Available: \url{https://github.com/RobotecAI/ros2-for-unity}
\BIBentrySTDinterwordspacing

\bibitem{attia2018global}
A.~Attia and S.~Dayan, ``Global overview of imitation learning,'' 2018.

\bibitem{bojarski2016end}
M.~Bojarski, D.~Del~Testa, D.~Dworakowski, B.~Firner, B.~Flepp, P.~Goyal, L.~D. Jackel, M.~Monfort, U.~Muller, J.~Zhang \emph{et~al.}, ``End to end learning for self-driving cars,'' \emph{arXiv preprint arXiv:1604.07316}, 2016.

\bibitem{sallab2017deep}
A.~E. Sallab, M.~Abdou, E.~Perot, and S.~Yogamani, ``Deep reinforcement learning framework for autonomous driving,'' \emph{arXiv preprint arXiv:1704.02532}, 2017.

\bibitem{kiran2021deep}
B.~R. Kiran, I.~Sobh, V.~Talpaert, P.~Mannion, A.~A.~A. Sallab, S.~Yogamani, and P.~Pérez, ``Deep reinforcement learning for autonomous driving: A survey,'' 2021.

\bibitem{ye2019deep}
H.~Ye, G.~Y. Li, and B.-H.~F. Juang, ``Deep reinforcement learning based resource allocation for v2v communications,'' \emph{IEEE Transactions on Vehicular Technology}, vol.~68, no.~4, pp. 3163--3173, 2019.

\end{thebibliography}

\end{document}